\documentclass{article}

\usepackage{arxiv}

\usepackage[utf8]{inputenc} % allow utf-8 input
\usepackage[T1]{fontenc}    % use 8-bit T1 fonts
\usepackage{hyperref}       % hyperlinks
\usepackage{url}            % simple URL typesetting
\usepackage{booktabs}       % professional-quality tables
\usepackage{amsfonts}       % blackboard math symbols
\usepackage{nicefrac}       % compact symbols for 1/2, etc.
\usepackage{microtype}      % microtypography
\usepackage{lipsum}		% Can be removed after putting your text content
\usepackage{graphicx}
\usepackage{natbib}
\usepackage{doi}
\usepackage{balance} % for balancing columns on the final page
\usepackage{amsmath}
\usepackage{algorithm}
\usepackage{algorithmic}
\usepackage{mathrsfs}
\usepackage{bm}
\usepackage{booktabs}
\usepackage{amsfonts,amssymb}
\usepackage{colortbl}
\usepackage{color}
\usepackage{tikz}
\usepackage{multicol}
\usepackage{threeparttable}
\usepackage{makecell}
\usepackage{multirow}
\title{Preference Inference from Demonstration in Multi-objective Multi-agent Decision Making}

\author{ \href{https://orcid.org/0000-0002-6014-9419}{\includegraphics[scale=0.06]{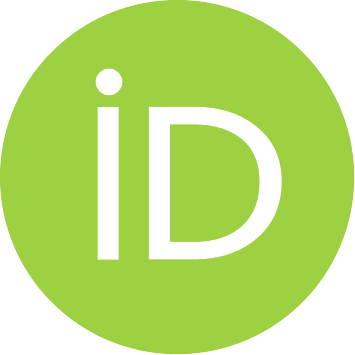}\hspace{1mm}Junlin Lu}
\thanks{Junlin Lu's research is supported by the Government of Ireland Postgraduate Scholarship (GOIPG/2022/2140).} 
\\
	School of Computer Science\\
	University of Galway\\
	\texttt{J.Lu5@nuigalway.ie} \\
}

\renewcommand{\shorttitle}{Preference Inference from Demonstration in Multi-objective Multi-agent Decision Making}

\hypersetup{
pdftitle={Preference Inference from Demonstration in Multi-objective Multi-agent Decision Making},
pdfsubject={cs.AI},
pdfauthor={Junlin Lu},
pdfkeywords={First keyword, Second keyword, More},
}

\begin{document}
\maketitle

\begin{abstract}
It is challenging to quantify numerical preferences for different objectives in a multi-objective decision-making problem. However, the demonstrations of a user are often accessible. We propose an algorithm to infer linear preference weights from either optimal or near-optimal demonstrations. The algorithm is evaluated in three environments with two baseline methods. Empirical results demonstrate significant improvements compared to the baseline algorithms, in terms of both time requirements and accuracy of the inferred preferences. In future work, we plan to evaluate the algorithm's effectiveness in a multi-agent system, where one of the agents is enabled to infer the preferences of an opponent using our preference inference algorithm.
\end{abstract}

% keywords can be removed

\section{Introduction}
In a multi-objective decision-making process, the agent receives reward vectors for different objectives. The utility function is used to make trade-offs between competing objectives by evaluating the reward vector as a utility scalar. In the existing literature, the most frequently used approach to get utility scalar is linear scalarization \citep{castelletti2013multiobjective,khamis2014adaptive,ferreira2017multi,lu2022multi}. In linear scalarization, the weight over the reward is referred to as the \textit{preference}. However, giving a precise numerical preference is not always intuitive for users. For example, consider a case where a portfolio manager selects stocks based on their weighting of minimizing risk and maximizing profits. He/she might want to give a higher weighting to maximize profits but a specific value is hard to determine. A small error in their preference can result in a significantly different policy which may lead to a sub-optimal solution. However, although it is difficult to numerically name the preference, a user can often demonstrate their preferences. Preference Inference (PI) methods that use demonstration are therefore helpful when solving such problems. 

Furthermore, in a multi-agent multi-objective decision-making process, if some agent can use the PI mechanism to infer other agents' preferences, it could gain information about the other agents. The knowledge of others' preferences can provide an advantage to these "wise" agents.

%%%%%%%%%%%%%%%%%%%%%%%%%%%%%%%%%%%%%%%%%%%%%%%%%%%%%%%%%%%%%%%%%%%%%%%%

\section{Preference Inference}
 PI is to infer a set of weights that are used to scalarize a reward vector during the multi-objective decision-making process. This is similar to inverse reinforcement learning (IRL), which infers the reward function by finding a set of linear parameters that scalarize state-relevant features. There are two assumptions:
\begin{itemize}
    \item The trajectories observed are from the optimal policy or near-optimal policy. This is a widely accepted assumption in existing literature for both IRL \citep{ng2000algorithms,ziebart2008maximum,wulfmeier2015deep} and PI \citep{takayama2022multi}.
    \item Based on the first assumption, given either optimal or near-optimal policy, the average reward trajectory is solely determined by preferences and environment transitions. 
\end{itemize}

The PI problem happens when we are given a point assumed to be on the Pareto optimal set (POS) based on some unknown weights for the utility function, and we would like to know what exactly the weights are. For more realistic scenarios, we also consider dominated points that are close to the points on the POS, known as \textit{sub-optimal policies} to test the PI algorithm. By adding sub-optimal noise to the data, we ensure that the inference model is robust on sub-optimal reward trajectories.

\begin{figure}[h]
    \centering
    \includegraphics[width=12cm]{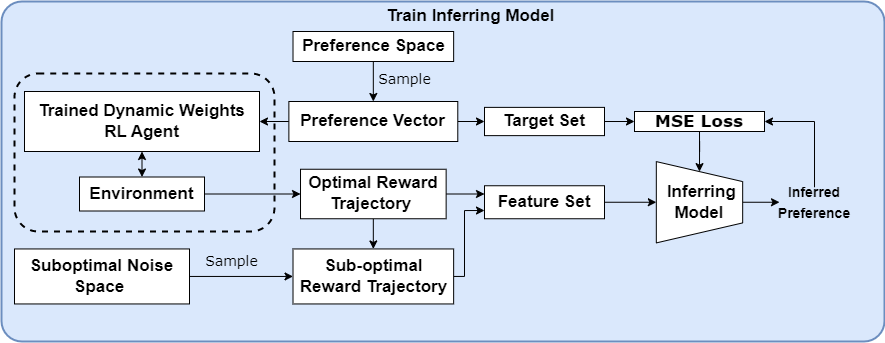}
    \caption{DWPI Training Phase}
    \label{fig:Preference Elicitator Training}
\end{figure}

We first use the dynamic weight reinforcement learning (DWRL) agent to generate behaviour trajectories \citep{kallstrom2019tunable}. We then propose the dynamic weight-based preference inference (DWPI) algorithm to infer the preferences of the agent for different behaviour trajectories. The training process of the DWPI algorithm is presented in Figure \ref{fig:Preference Elicitator Training}.
The DWRL agent takes the preference vector as part of the state so that it can change its behavior pattern during runtime by changing the preference. Given the interaction between a trained DWRL agent and the environment, a set of optimal reward trajectories are generated. The trajectory set is augmented by adding random sampled sub-optimal noise to be the training set of the inference model. The inference model is trained under the supervised learning paradigm by inputting the reward trajectory and predicting the corresponding preference. 
\begin{algorithm}[h]\small
\caption{Dynamic Weight Preference Inference Algorithm}
\label{alg:DWPIAlgorithm}
\begin{algorithmic}
\STATE {Initialize inferring model $\mathcal{I}$, sub-optimal noise space $\mathcal{SN}$, environment $\mathcal{E}$, and preference space $\Omega$}
\STATE {Load the trained dynamic weights RL agent $AG$}
\STATE {Initialize feature set $X$, target set $Y$}
\WHILE {not enough entries in $X$}
    \STATE {Sample a preference vectors $\bm{\omega}$ from $\Omega$}
    \STATE {$AG$ plays one episode with $\bm{\omega}$, generates reward trajectory $\bm{\tau_{r}}$}
    \STATE {Sample a noise vector $\bm{\delta}$ from $\mathcal{SN}$}
    \STATE {Store the noisy reward trajectory $\bm{\tau_{r}}+\bm{\delta}$ in $X$, store $\bm{\omega}$ in $Y$}
\ENDWHILE
\WHILE{$\mathcal{I}$ not converge}
\STATE {Sample batch from $X$ to train the inferring model, loss $\mathcal{L}=\lVert\hat{\bm{\omega}}-\bm{\omega}\rVert$}
\ENDWHILE
\end{algorithmic}
\end{algorithm}

\begin{figure}[h]
        \centering
        \includegraphics[width=12cm]{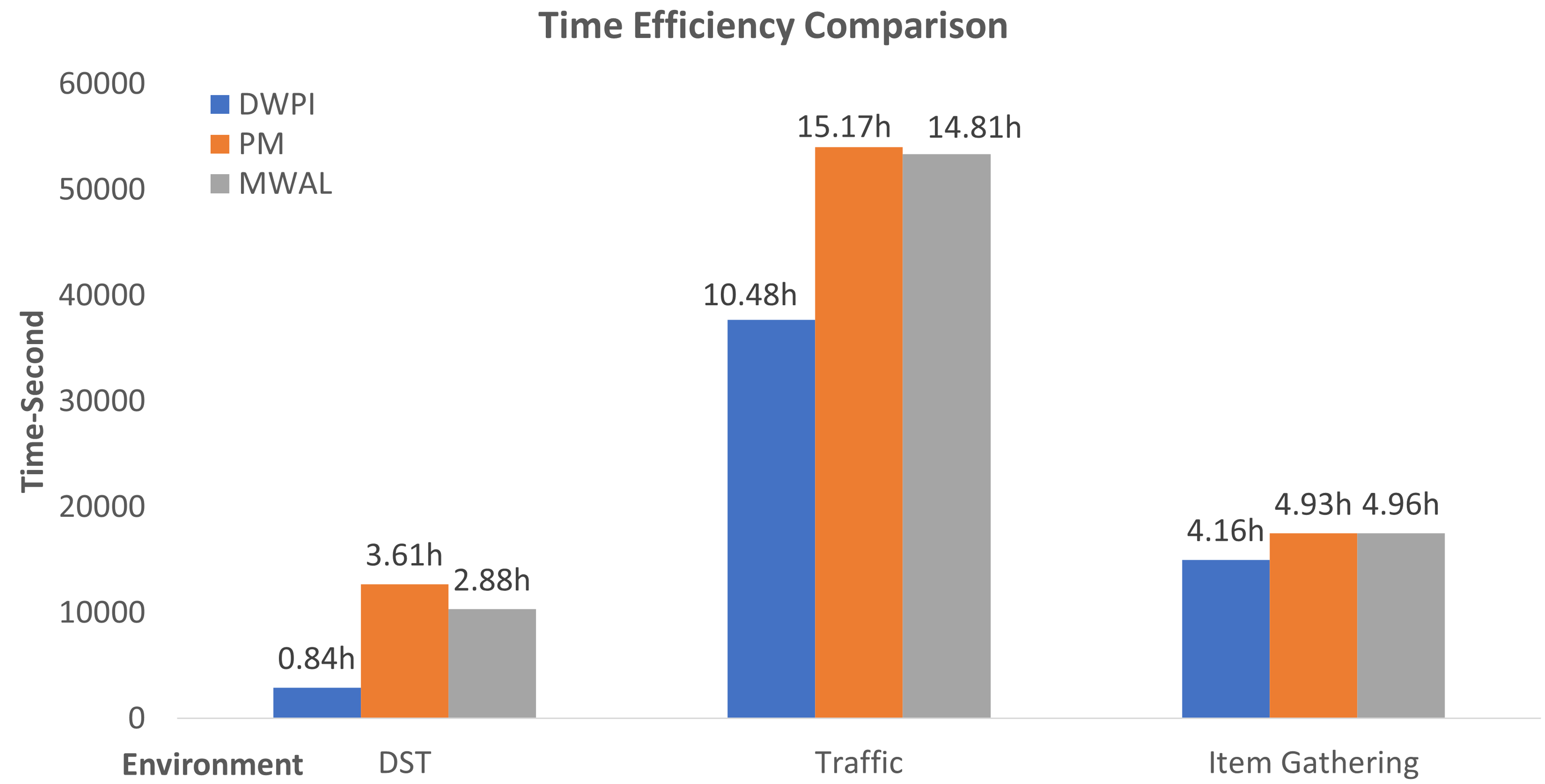}
        \caption{Time Efficiency Comparison}
        \label{fig:Time Efficiency Comparison}
\end{figure}

We evaluate our algorithm in three environments: Convex Deep Sea Treasure\citep{mannion2017policy}, Traffic\citep{kallstrom2019tunable}, and Item Gathering\citep{kallstrom2019tunable} and compared to two benchmarks projection method (PM) 
\citep{ikenaga2018inverse} and multiplicative weights apprenticeship learning (MWAL) \citep{takayama2022multi}. 
Our method outperforms the baseline methods by both time efficiency (Figure \ref{fig:Time Efficiency Comparison}) and PI performance (Table \ref{tab:Performance Improvement}). The experiments are implemented with Python 3.9, TensorFlow version 2.3.0, and run on a machine with 11th Gen Intel(R) Core(TM) i7-1165G7 \@ 2.80GHz CPU.
\begin{table}[h]\small
\centering
\caption{Performance Improvement}
\label{tab:Performance Improvement} 
\begin{tabular}{c|c|c|c|c|c}
    \toprule
    \multicolumn{2}{c}{Environment}&\multicolumn{2}{|c}{Traffic}&\multicolumn{2}{|c}{Item Gathering}\\
    \midrule
    \textbf{KL-divergence}&&PM&MWAL&PM&MWAL\\
    \midrule
    Optimal Demo&DWPI&90.8\%$\uparrow$&99.56\%$\uparrow$&98.01\%$\uparrow$&99.13\%$\uparrow$\\
    Sub-optimal Demo&DWPI&89.55\%$\uparrow$&99.53\%$\uparrow$&96.89\%$\uparrow$&99.25\%$\uparrow$\\
    \midrule
    \textbf{Mean Squared Error}&&PM&MWAL&PM&MWAL\\
    \midrule
    Optimal Demo&DWPI&97.7\%$\uparrow$&99.8\%$\uparrow$&85.91\%$\uparrow$&98.1\%$\uparrow$\\
    Sub-optimal Demo&DWPI&94.13\%$\uparrow$&99.83\%$\uparrow$&83.81\%$\uparrow$&98.9\%$\uparrow$\\
    \midrule
    \textbf{Utility}&&PM&MWAL&PM&MWAL\\
    \midrule
    Optimal Demo&DWPI&60.56\%$\uparrow$&98.93\%$\uparrow$&90.67\%$\uparrow$&82.62\%$\uparrow$\\
    Sub-optimal Demo&DWPI&71.87\%$\uparrow$&90.60\%$\uparrow$&99.85\%$\uparrow$&94.50\%$\uparrow$\\
    \bottomrule
\end{tabular}
\end{table}

The results of the evaluation show that the DWPI algorithm performs well in terms of inference accuracy. 
\section{Opponent Preference Modelling}
 In future work, we will utilize the DWPI algorithm in the multi-agent environment to enable an agent to gain knowledge of other agents' preferences to gain an advantage over them. 
 
 It will be evaluated in two environments. The first one is the Wolfpack environment \citep{leibo2017multi}, where two predator agents try to capture randomly moving prey. If the capture happens when the Manhattan distance between the two predators $dist\leq3$, it is determined as cooperation or competition when $dist>3$. 
 
 The second environment is the multi-agent item gathering modified from \citep{kallstrom2019tunable}. Two RL agents move in the grid world to gather randomly distributed blocks with three different colors, i.e. green, red, and yellow. Each agent has a preference weight vector over the color of the blocks.
 
Two sets of experiments will be done in each environment. The first experiment is between two normal agents while the second experiment is between a normal agent and a wiser agent which is able to infer the other's preference. The performance of the agents will be analyzed to check whether the inference mechanism will help the wiser agent gain advantages during the games.

\section{Conclusion}
We propose the DWPI algorithm to infer preference from demonstrations in the multi-objective decision-making process. We further evaluate the utility of this algorithm for enhancing an agent in the multi-agent system. With our PI model, an agent can know its opponent better and therefore achieve better performance on its target. For future work, we would like to evaluate the opponent preference modeling performance in multi-agent systems. The extension of the DWPI algorithm to infer a non-linear preference is also a direction that we are interested in. 
%%%%%%%%%%%%%%%%%%%%%%%%%%%%%%%%%%%%%%%%%%%%%%%%%%%%%%%%%%%%%%%%%%%%%%%%

%%%%%%%%%%%%%%%%%%%%%%%%%%%%%%%%%%%%%%%%%%%%%%%%%%%%%%%%%%%%%%%%%%%%%%%%

%%% The acknowledgments section is defined using the "acks" environment
%%% (rather than an unnumbered section). The use of this environment 
%%% ensures the proper identification of the section in the article 
%%% metadata as well as the consistent spelling of the heading.

\bibliographystyle{unsrtnat}
\bibliography{references}  %%% Uncomment this line and comment out the ``thebibliography'' section below to use the external .bib file (using bibtex) .





% \end{thebibliography}

\end{document}